\def\year{2019}\relax
\documentclass[letterpaper]{article} 
\usepackage{aaai19}  
\usepackage{times}  
\usepackage{helvet}  
\usepackage{courier}  
\usepackage{url}  
\usepackage{graphicx}  

\usepackage{subcaption}
\usepackage{multirow}
\usepackage{framed}
\usepackage{mathtools}

\newtheorem{proof}{PROOF}
 \newtheorem{definition}{Definition}
 
\newtheorem{lemma}{Lemma}

\DeclareMathOperator*{\argmin}{argmin}

\usepackage{amsmath,amssymb,amsfonts}
\usepackage{algorithm}
\usepackage{algorithmic}

\usepackage{xcolor}
\definecolor{DSgray}{cmyk}{0,1,0,0}

\begin{document}

\title{Nearest-Neighbour-Induced Isolation Similarity \\ and Its Impact on Density-Based Clustering}

 \author{Xiaoyu Qin\\
{Monash University}\\
Victoria, Australia 3800\\
xiaoyu.qin@ieee.org
\And
Kai Ming Ting\\
{Federation University}\\
Victoria, Australia 3842\\
kaiming.ting@federation.edu.au
\And
Ye Zhu\\
{Deakin University}\\
Victoria, Australia 3125\\
ye.zhu@ieee.org
\And
Vincent CS Lee\\
{Monash University}\\
Victoria, Australia 3800\\
vincent.cs.lee@monash.edu}

\maketitle

\begin{abstract}
A recent proposal of data dependent similarity called Isolation Kernel/Similarity has enabled SVM to produce better classification accuracy. We identify shortcomings of using a tree method to implement Isolation Similarity; and propose a nearest neighbour method instead. We formally prove the characteristic of Isolation Similarity with the use of the proposed method.  
The impact of Isolation Similarity on density-based clustering is studied here.
We show for the first time that the clustering performance of the classic density-based clustering algorithm DBSCAN can be significantly uplifted to surpass that of the recent density-peak clustering algorithm DP.
This is achieved by simply replacing the distance measure with the proposed nearest-neighbour-induced Isolation Similarity in DBSCAN, leaving the rest of the procedure unchanged.
A new type of clusters called mass-connected clusters is formally defined. We show that DBSCAN, which detects density-connected clusters, becomes one which detects mass-connected clusters, when the distance measure is replaced with the proposed similarity. We also provide the condition under which mass-connected clusters can be detected, while density-connected clusters cannot.
\end{abstract}

\section{Introduction}
Similarity measure is widely used in various data mining and machine learning tasks. In clustering analysis, its impact to the quality of result is critical\cite{steinbach2004challenges}. 
A recent proposal of data dependent similarity called Isolation Kernel has enabled SVM to produce better classification accuracy by simply replacing the commonly used data independent kernel (such as Gaussian kernel) with Isolation Kernel \cite{ting2018IsolationKernel}. This is made possible on datasets of varied densities because Isolation Kernel is adaptive to local data distribution such that two points in a sparse region are more similar than two points of equal inter-point distance in a dense region.
Despite this success, the kernel characteristic has not been formally proven yet.

This paper extends this line of research by investigating a different implementation of Isolation Similarity. We provide a formal proof of the characteristic of the Isolation Similarity for the first time since its introduction. In addition, we focus on using Isolation Similarity to improve the clustering performance of density-based clustering.

This paper identifies shortcomings of tree-based method currently employed in inducing Isolation Similarities \cite{ting2018IsolationKernel}. Instead, we investigate a different method to induce the data dependent Isolation Similarity, and evaluate its clustering performance using DBSCAN \cite{ester1996density} in comparison with the two existing improvements of density-based clustering, i.e., DScale \cite{DSCALE} and DP \cite{rodriguez2014clustering}. 

The rest of the paper is organised as follows. We reiterate Isolation Kernel, identify the shortcomings of using the tree-based method to induce Isolation Similarity, provide the proposed alternative that employs a nearest neighbour method,  the lemma and proof of the characteristic of Isolation Similarity, and the investigation in using Isolation Similarity in density-based clustering.

The descriptions of existing works are framed in order to clearly differentiate  from the contributions we made here.

\section{Isolation Kernel}
\label{sec_Isolation_Kernel}
Isolation Kernel/Similarity is first proposed by \cite{ting2018IsolationKernel} as a new similarity which can adapt to density structure of the given dataset, as opposed to commonly used data independent kernels such as Gaussian and Laplacian kernels.

In the classification context, Isolation Kernel has been shown to be an effective means to improve the accuracy of SVM, especially in datasets which have varied densities in the class overlap regions \cite{ting2018IsolationKernel}. This is achieved by simply replacing the commonly used data independent kernel such as Gaussian and Laplacian kernels with the Isolation Kernel. 

In the context of SVM classifiers, Isolation Kernel \cite{ting2018IsolationKernel} has been shown to be more effective than existing approaches such as distance metric learning \cite{zadeh2016geometric,Wang2015}, multiple kernel learning \cite{rakotomamonjy2008simplemkl,MKL2011} and Random Forest kernel \cite{Breiman2000,Davis2014}.

The characteristic of Isolation Kernel is akin to one aspect of human-judged similarity as discovered by psychologists \cite{Krumhansl,Tversky}, i.e., human will judge the two same Caucasians as less similar when compared in Europe (which have many Caucasians) than in Asia.

We restate the definition and kernel characteristic \cite{ting2018IsolationKernel} below.

\begin{framed}
Let $D=\{x_1,\dots,x_n\}, x_i \in \mathbb{R}^d$ be a dataset sampled from an unknown probability density function $x_i \sim F$.
Let $\mathcal{H}_\psi(D)$ denote the set of
all partitions $H$
that are admissible under $D$
where each isolating partition $\theta \in H$ isolates one data point from the rest of the points in a random subset $\mathcal D \subset D$, and $|\mathcal D|=\psi$.

\begin{definition}
For any two points $x, y \in \mathbb{R}^d$,
Isolation Kernel of $x$ and $y$ wrt $D$ is defined to be
the expectation taken over the probability distribution on all partitioning $H \in {\mathcal H}_\psi(D)$ that both $x$ and $y$ fall into the same isolating partition $\theta \in H$:
\begin{equation}
K_\psi(x,y|D) =  {\mathbb E}_{{\mathcal H}_\psi(D)} [\mathbb{I}(x,y \in \theta\ | \ \theta \in H)]
\label{eqn_kernel}
\end{equation}

\noindent
where $\mathbb{I}(B)$ is the indicator function which outputs 1 if $B$ is true; otherwise, $\mathbb{I}(B)=0$.
\label{def_Isolation_Kernel}
\end{definition}

In practice, $K_\psi$ is estimated from a finite number of partitionings $H_i \in \mathcal{H}_\psi(D), i=1,\dots,t$ as follows:
\begin{eqnarray}
K_\psi(x,y|D) =  \frac{1}{t} \sum_{i=1}^t   \mathbb{I}(x,y \in \theta\ | \ \theta \in H_i)
\label{eqn_kernel2}
\end{eqnarray}

The characteristic of Isolation Kernel is: {\bf two points in a sparse region are more similar than two points of equal inter-point distance in a dense region}, i.e.,

\vspace{2mm}
{\bf Characteristic of $K_\psi$}:
$\forall x, y \in \mathcal{X}_\mathsf{S}$ and $\forall x',y' \in \mathcal{X}_\mathsf{T}$ such that $\parallel x-y \parallel\ =\ \parallel x'- y'\parallel$,
$K_\psi$ satisfies the following condition:
\begin{eqnarray}
K_\psi( x, y) >    K_\psi( x', y')
\label{eqn_condition}
\end{eqnarray}

\noindent
where $\mathcal{X}_\mathsf{S}$ and $\mathcal{X}_\mathsf{T}$ are two subsets of points in sparse and dense regions of $\mathbb{R}^d$, respectively;
and $\parallel x-y\parallel$ is the distance between $x$ and $y$.

To get the above characteristic, the required
property of the space partitioning mechanism is to create large partitions in the sparse region and small partitions in the dense region such that \emph{two points are more likely to fall into a same partition in a sparse region than two points of equal inter-point distance in a dense region}. 
\end{framed}

\section{Shortcomings of\\ tree-based isolation partitioning}
\label{sec_shortcomings}
Isolation Kernel \cite{ting2018IsolationKernel} employs isolation trees or iForest \cite{liu2008isolation} to measure the similarity of two points because its space partitioning mechanism produces the required partitions which have volumes that are monotonically decreasing wrt the density of the local region.

Here we identify two shortcomings in using isolation trees to measure Isolation Similarity, i.e., each isolation tree (i) employs axis-parallel splits; and (ii) is an imbalanced tree.

\begin{figure}
\centering
\begin{subfigure}{0.23\textwidth}
  \centering
  \includegraphics[width=\textwidth]{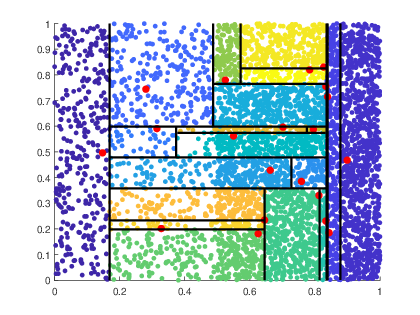}
  \caption{Axis-parallel splitting}
\end{subfigure}
\begin{subfigure}{0.23\textwidth}
  \centering
  \includegraphics[width=\textwidth]{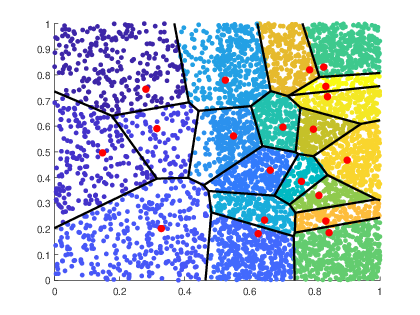}
  \caption{NN partitioning}
\end{subfigure}
\caption{Examples of two isolation partitioning mechanisms: Axis-parallel  versus nearest neighbour (NN). On a dataset having two (uniform) densities, i.e., the right half has a higher density than the left half.}
\label{fig_compare}
\end{figure}

Figure \ref{fig_compare}(a) shows an example partitioning due to axis-parallel splits of an isolation tree.
The tree-based isolating partitions generally satisfy the requirement of small partitions in dense region and large partitions in sparse region. 

However, it produced some undesirable effect, i.e., some partitions are always overextended for the first few splits close to the root of an imbalanced tree\footnote{Imbalanced trees are a necessary characteristic of isolation trees \cite{liu2008isolation} for their intended purpose of detecting anomalies, where anomalies are expected to be isolated with few splits; and normal points can only be isolated using a large number of splits.}. These are manifested as elongated rectangles in Figure \ref{fig_compare}(a).

While using balanced trees can be expected to overcome this problem, the restriction to hyper-rectangles remains due to the use of axis-parallel splits.

To overcome these shortcomings of isolation trees, we propose to use a nearest neighbour partitioning mechanism which creates a Voronoi diagram \cite{aurenhammer1991voronoi} where each cell is an isolating partition (i.e., isolating one point from the rest of the points in the given sample.) An example is provided in Figure \ref{fig_compare}(b). Note that these partitions also satisfy the requirement of small partitions in the dense region and large partitions in the sparse region. But they do not have the undesirable effect of elongated rectangles.
We provide our implementation in the next section.

\section{Nearest neighbour-induced\\ Isolation Similarity}
\label{sec_NN_Method}
Instead of using trees in its first implementation \cite{ting2018IsolationKernel}, we propose to implement Isolation Similarity using nearest neighbours. 

Like the tree method, the nearest neighbour method also produces each $H$ model which consists of $\psi$ isolating partitions $\theta$, given a subsample of $\psi$ points. Rather than representing each isolating partition as a hyper-rectangle, it is represented as a cell in a Voronoi diagram \cite{aurenhammer1991voronoi}, where the boundary between two points is the equal distance from these two points.

While the Voronoi diagram is nothing new, its use in measuring similarity is new.

Using the same notations as used earlier, $H$ is now a Voronoi diagram, built by employing $\psi$ points in $\mathcal D$,
where each isolating partition or Voronoi cell $\theta \in H$ isolates one data point from the rest of the points in $\mathcal D$. We call the point which determines a cell as the cell centre.

Given a Voronoi diagram $H$ constructed from a sample $\mathcal{D}$ of $\psi$ points, the Voronoi cell centred at $z \in \mathcal{D}$ is:
\[ \theta[z] = \{x  \in \mathbb{R}^d \ | \  z = \argmin_{\mathsf{z} \in \mathcal{D}} \ell_p(x - \mathsf{z})\}. \]
\noindent where $\ell_p(x, y)$ is a distance function and we use $p=2$ as Euclidean distance in this paper. 

\label{sec_proof}

\begin{definition}
For any two points $x, y \in \mathbb{R}^d$, the nearest neighbour-induced
Isolation Similarity of $x$ and $y$ wrt $D$ is defined to be
the expectation taken over the probability distribution on all Voronoi diagrams $H \in {\mathcal H}_\psi(D)$ that both $x$ and $y$ fall into the same Voronoi cell $\theta \in H$:

\begin{eqnarray}
K_\psi(x,y\ |\ D) &=&  {\mathbb E}_{{\mathcal H}_\psi(D)} [\mathbb{I}(x,y \in \theta[z]\ | \ \theta[z] \in H)] \nonumber \\
&=& {\mathbb E}_{\mathcal{D} \sim D} [\mathbb{I}(x,y\in \theta[z]\ | \ z\in \mathcal{D})]  \nonumber
\\
&=& P(x,y\in \theta[z]\ | \ z\in \mathcal{D} \subset D)
\label{eqn_kernel_anne}
\end{eqnarray}
\noindent
where  $P$ denotes the probability.
\label{def_anne_similarity}
\end{definition}

The Voronoi diagram has the required property of the space partitioning mechanism to produce large partitions in a sparse region and small partitions in a dense region. This yields the characteristic of Isolation Similarity : {\bf two points in a sparse region are more similar than two points of equal inter-point distance in a dense region}. 

The use of nearest neighbours facilitates a proof of the above characteristic that was previously hampered by the use of trees.
We provide the proof in the next section.

\section{Lemma and Proof of the characteristic of Isolation Similarity}

Let $\rho(x)$ denote the density at point $x$, a lemma based on definition 4 is given below:

\begin{lemma}
$\forall x, y \in \mathcal{X}_\mathsf{S}$ (sparse region) and $\forall x',y' \in \mathcal{X}_\mathsf{T}$ (dense region) such that $\forall_{z\in \mathcal{X}_\mathsf{S}, z'\in \mathcal{X}_\mathsf{T}} \ \rho(z)<\rho(z')$,
the nearest neighbour-induced Isolation Similarity $K_\psi$ has the characteristic that for $\ell_p(x-y)\ =\ \ell_p(x'- y')$ implies
\begin{eqnarray}
 P(x,y\in \theta[z]) > P(x',y'\in \theta[z'])  \equiv \hspace{3cm} \nonumber \\ K_\psi( x, y\ |\ D) > K_\psi( x', y'\ |\ D) \nonumber
\label{eqn_condition0}
\end{eqnarray}
\end{lemma}

Sketch of the proof: (i) If two points fall into the same Voronoi cell, then the distances of these individual points to this cell centre must be shorter than those to every other cell centre (or at most equal to those to one other cell centre) in a Voronoi diagram formed by all these cell centres. (ii) In a subset of $\psi$ points, sampled from $D$, used to form a Voronoi diagram, the probability of two points falling into the same Voronoi cell can then be estimated based on the condition stated in (i). (iii) The probability of two points of equal inter-point distance falling into the same Voronoi cell is a monotonically decreasing function wrt the density of the cell.  

\begin{proof}
Let a local region $V(x,y)$ covering both $x$ and $y$  as a ball centred at the middle between $x$ and $y$ having $\ell_p(x,y)$ as the diameter of the ball.
Assume that the density in $V(x,y)$ is uniform and denoted as  $\rho(V(x,y))$.

Let $\mathcal{N}_\epsilon(x)$ be the $\epsilon$-neighbourhood of $x$, i.e., $\mathcal{N}_\epsilon(x)=\lbrace y \in D ~|~ \ell_p(x,y) \leqslant \epsilon \rbrace$. The probability of both $x$ and $y$ are in the same Voronoi cell $\theta [z]$ is equivalent to the probability of a point $z\in \mathcal{D}$ being the nearest neighbour of both $x$ and $y$ wrt all other points in $\mathcal{D}$, i.e., the probability of selecting $\psi-1$ points which are all located outside the region $U(x,y,z)$, where $U(x,y,z)=\mathcal{N}_{\ell_p(x,z)}(x) \cup \mathcal{N}_{\ell_p(y,z)}(y)$.

To simplify notation, $z\in \mathcal{D}$ is omitted. Then the probability of  $x,y\in \theta[z]$ can be expressed as follows:
\begin{eqnarray} 
P(x,y\in \theta[z]\ | \ z\in V(x,y))
\mbox{\hspace{3cm}} \nonumber\\ 
= P(z_1,z_2,\dots,z_{(\psi-1)} \notin U(x,y,z)) \hspace{7mm}
\nonumber\\  \propto (1-{\mathbb E}_{z \sim V(x,y)} [|U(x,y,z)|]/|D|)^{(\psi-1)} \nonumber
\label{pro2}
\end{eqnarray}

\noindent
where $|W|$ denotes the cardinality of $W$.

Assume that $U(x,y,z)$ is also uniformly distributed, having the same density $\rho(V(x,y))$, the expected value of $|U(x,y,z)|$ can be estimated as:
\begin{eqnarray} 
{\mathbb E}_{z  \sim  V(x,y)} [|U(x,y,z)|] \mbox{\hspace{4cm}} \nonumber \\
= {\mathbb E}_{z  \sim  V(x,y)} [\upsilon(U(x,y,z)) \times \rho(V(x,y))] \nonumber\\
= {\mathbb E}_{z  \sim  V(x,y)} [\upsilon(U(x,y,z))] \times \rho(V(x,y)) \nonumber
\label{pro4}
\end{eqnarray}

\noindent where $\upsilon(W)$ denotes the volume of $W$.

Thus, we have 
\begin{eqnarray} 
P(x,y\in \theta[z]\ | \ z\in V(x,y))
\propto \mbox{\hspace{3cm}}  \nonumber\\
\Big(1-{\mathbb E}_{z  \sim  V(x,y)} [\upsilon(U(x,y,z))]\times  \frac{\rho(V(x,y))}{|D|}\Big)^{(\psi-1)}
\label{pro5}
\end{eqnarray}

In other words, the higher the density in the area around $x$ and $y$,
the smaller $P(x,y\in \theta[z]\ | \ z\in V(x,y))$
is, as the volume of $V(x,y)$ is  constant given $x$ and $y$. 

Given two pairs of points from two different regions but of equal interpoint distance as follows: 
$\forall x, y \in \mathcal{X}_\mathsf{S}$ (sparse region) and $\forall x',y' \in \mathcal{X}_\mathsf{T}$ (dense region) such that  $\ell_p(x, y)\ =\ \ell_p(x', y')$.
 
Assume that data are uniformly distributed in both regions, and we sample $z,z' \in \mathcal{D}$ from $D$ such that $z\in V(x,y)$ and $z'\in V(x',y')$. We have ${\mathbb E}_{z \sim V(x,y)} [\upsilon(U(x,y,z))]={\mathbb E}_{z' \sim V(x',y')} [\upsilon(U(x',y',z'))]$ because the volume of $V(x,y)$ is equal to that of $V(x',y')$ for $\ell_p(x, y)\ =\ \ell_p(x', y')$,
independent of the density of the region.

Supposing that we choose a sufficient large sample size $\psi$ of $\mathcal{D}$ which contains points from both $V(x,y)$ and $V(x',y')$. When the data are uniformly distributed in $U(x,y,z) \in \mathcal{X}_\mathsf{S}$ and $U(x',y',z') \in \mathcal{X}_\mathsf{T}$, based on Equation \ref{pro5}, we have
\begin{eqnarray} 
 P(x,y\in \theta[z]\ | \ z\in V(x,y))
 >  \hspace{3cm} \nonumber \\ P(x',y'\in \theta[z']\ | \ z'\in V(x',y'))
 \nonumber\\
 \equiv K_\psi(x,y\ |\ D) > K_\psi(x,y\ |\ D)  \hspace{3cm}\nonumber
\label{eq10}
\end{eqnarray} 

This means that $x'$ and $y'$ (in a dense region) are more like to be in different cells than $x$ and $y$ in $V(x,y)$ (in a sparse region),  as shown in Figure~\ref{fig_compare}.
\hfill $\square$\\ 
\end{proof}

\begin{figure}
\centering
\captionsetup{justification=centering}
\begin{subfigure}{0.4\textwidth}
\centering
  \includegraphics[width=100pt]{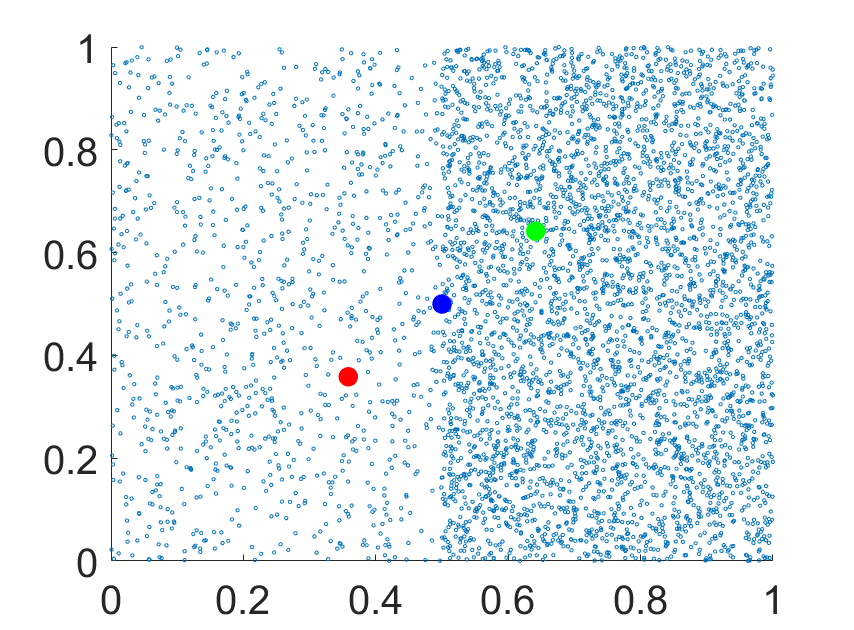}
  \caption{Two regions of different densities}
\label{gap:a}
\end{subfigure}\\
\begin{subfigure}{0.23\textwidth}
  \includegraphics[width=110pt]{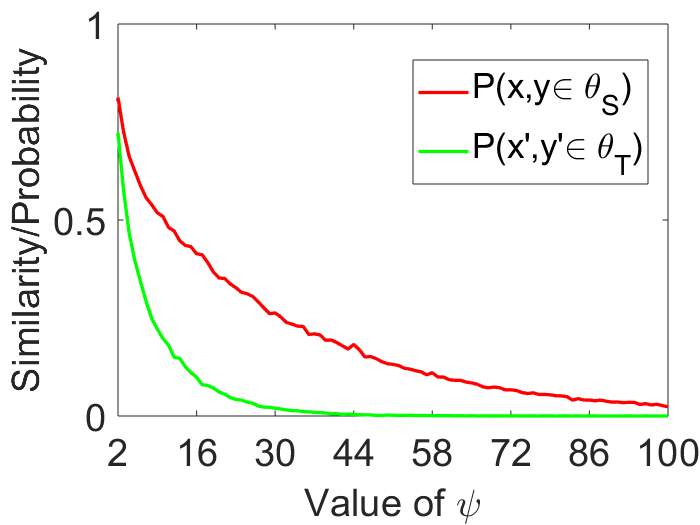}
  \caption{$\psi$ increases \\ Inter-point distance=0.2}
\label{gap:c}
\end{subfigure} 
\begin{subfigure}{0.23\textwidth}
  \includegraphics[width=110pt]{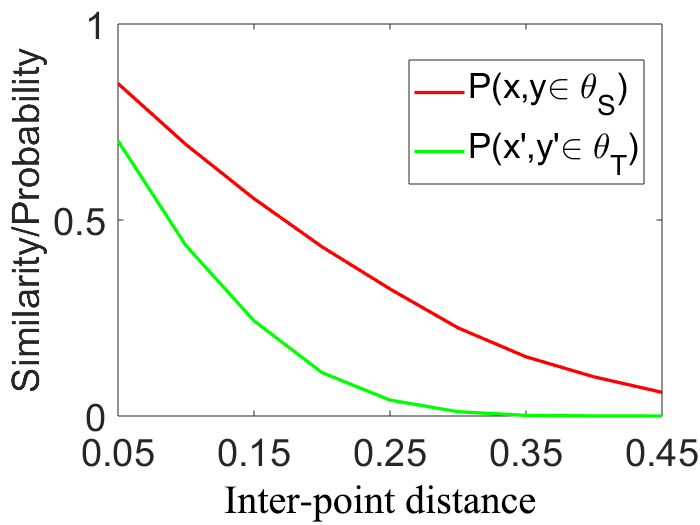}
  \caption{$\psi$=15 \\ Inter-point distance increases}
\label{gap:e}
\end{subfigure} 
\caption{(a) Reference points used in the simulations, where 
inter-point distance $\parallel x-y \parallel\ =\ \parallel x'- y'\parallel$ increases.
Simulation results  as $\psi$ increases (b);  and  as inter-point distance increases (c). $t=10000$ is used.}
\label{gap}
\end{figure}

\newpage
A simulation validating the above analysis is given in 
Figure~\ref{gap}. It compares $P(x,y\in \theta_\mathsf{S})$ and $P(x',y'\in \theta_\mathsf{T})$ when $x,y$ from a sparse region and $x',y'$ from a dense region with equal inter-point distance. Given a fixed $\psi < |D|$ or a fixed inter-point distance, properties observed from Figure~\ref{gap} are given as follows:

\begin{enumerate}
\item
$P(x,y\in \theta_\mathsf{S})>P(x',y'\in \theta_\mathsf{T})$.
\item
The rate of decrease of $P(x',y'\in \theta_\mathsf{T})$ is faster than that of $P(x,y\in \theta_\mathsf{S})$. Thus $P(x',y'\in \theta_\mathsf{T})$ reaches 0 earlier.
\end{enumerate}

\section{Isolation Dissimilarity and contour maps}
\label{sec_example_contour_map}
To be consistent with the concept of distance as a kind of dissimilarity, we use Isolation Dissimilarity hereafter:

Isolation dissimilarity: $\mathfrak p_\imath(x,y)= 1 - K_\psi(x,y)$.

Like $\ell_p$ norm, $\forall x, \mathfrak p_\imath(x,x) = 0$ and $\mathfrak p_\imath(x,y) = \mathfrak p_\imath(y,x)$. However, $\forall x \ne y,\  \mathfrak p_\imath(x,y)$ depends on the data distribution and how $\mathfrak p_\imath$ is implemented, not the geometric positions only.

\begin{figure}
\begin{subfigure}{0.23\textwidth}
  \centering
  \includegraphics[width=110pt,height=90pt]{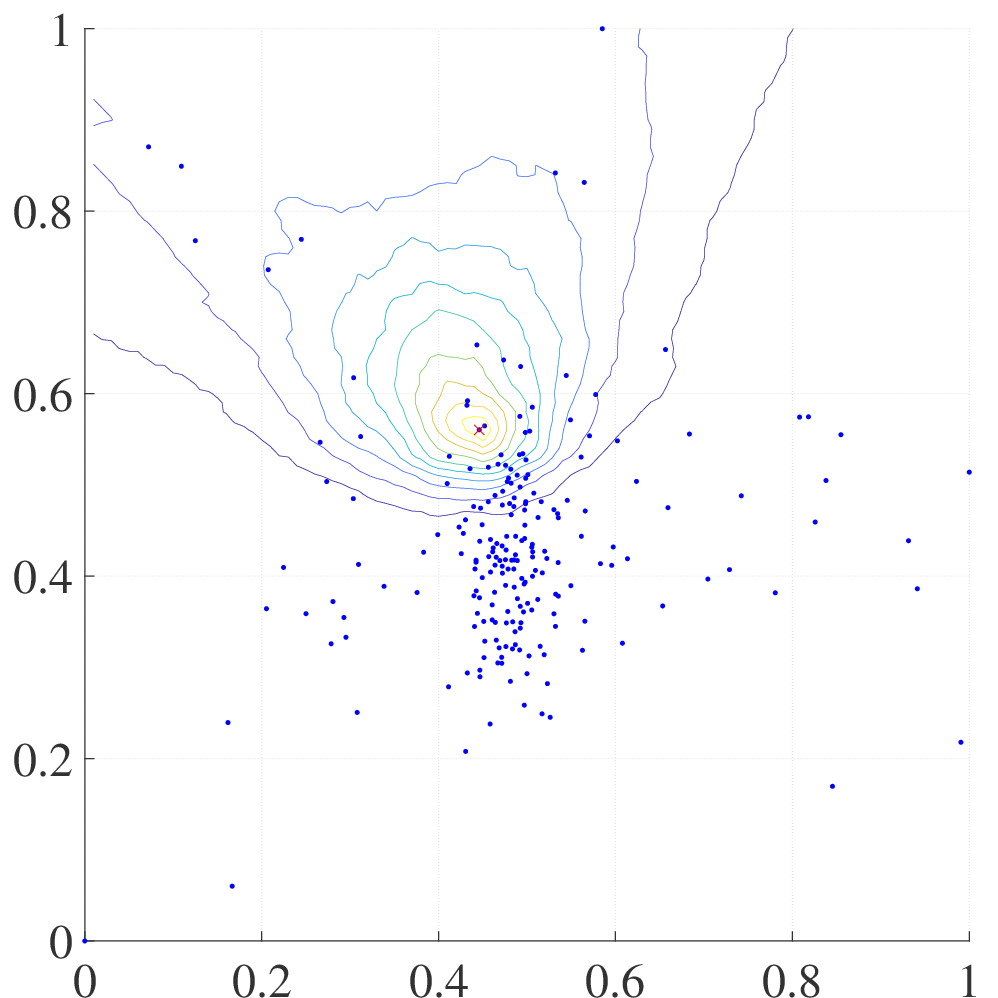}
  \caption{$\mathfrak p_\imath$-aNNE}
\end{subfigure}%
\begin{subfigure}{0.23\textwidth}
  \centering
  \includegraphics[width=110pt,height=90pt]{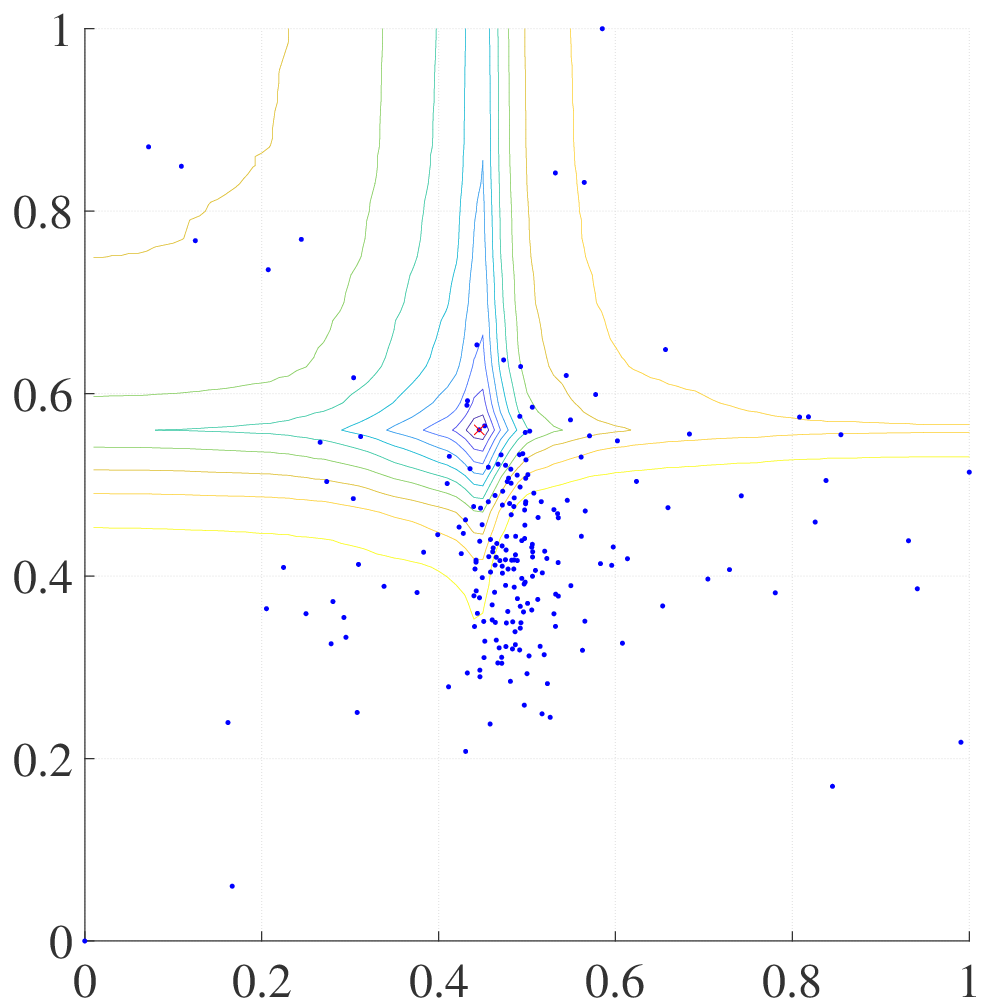}
  \caption{$\mathfrak p_\imath$-iForest}
\end{subfigure}%
\caption{Contour plots of 
$\mathfrak p_\imath$ on the Thyroid dataset (mapped to 2 dimensions using MDS \cite{borg2012applied}). $\psi=14$ is used in aNNE and iForest.}
\label{fig_contour_example}
\end{figure}

We denote the nearest-neighbour-induced Isolation Dissimilarity $\mathfrak p_\imath$-aNNE; and the tree-induced version $\mathfrak p_\imath$-iForest.
An example comparison of the contour maps produced the two dissimilarities are given in Figure~\ref{fig_contour_example}. Note that the contour maps of $\mathfrak p_\imath$ depend on the data distribution, whereas that of $\ell_2$  is not. Also, comparing to $\ell_2$, the other dissimilarity change slower in area far from the centre point and faster in area close to the centre point.

We examine the impact of Isolation Dissimilarity on density-based clustering in the rest of the paper. We describe the neighbourhood density function commonly used in the density-based clustering and its counterpart called neighbourhood mass function in the next section.

\section{Neighbourhood density and mass functions}
\label{sec_neighbourhood_function}

\begin{framed}
Neighbourhood mass function \cite{ting2016overcoming} was first proposed as a way to model data in terms of mass distribution, analogous to the neighbourhood density function used in modelling data as a density distribution.
The key difference is the measure $\partial$ used in the following  function of $x$: $\#\{y \in D\ |\ \partial(x,y) \le \mbox{cutoff} \}$, where the boundary of the region within which the points are counted is set by a user-specified constant cutoff. When a distance measure is used, it becomes a neighbourhood density function as the ball has a fixed volume when the cutoff is fixed. It is a neighbourhood mass function when a data dependent dissimilarity is used as the volume of the `ball' varies depending on the local distribution even if the cutoff is fixed.
\end{framed}

\section{Mass-connected clusters}
\label{sec_mass-connectivity}

Here we define mass-connected clusters  defined in terms of neighbourhood mass function: $$ M_\alpha(x) = \#\{ y \in D\ |\ \mathfrak p_\imath(x,y) \le \alpha\} $$

\begin{definition}
Using an $\alpha$-neighbourhood mass estimator $ M_\alpha(x) = \#\{ y \in D\ |\ \mathfrak p_\imath(x,y) \le \alpha\} $, mass-connectivity with  threshold $\tau$ between $x_1$ and $x_p$ via a sequence of $p$ unique points from $D$, i.e., $\{x_1,x_2,x_3,...,x_p\}$ is denoted as $MConnect_{\alpha}^{\tau}(x_1, x_p)$, and it is defined as:
 
\begin{equation}
\begin{split}
    MConnect_{\alpha}^{\tau}(x_1, x_p) & \leftrightarrow \\
    [(\mathfrak p_\imath(x_1,x_2)\leq \alpha)  \wedge & ((M_{\alpha}(x_1)\geq \tau)  \vee  (M_{\alpha}(x_2)\geq \tau))] \\ 
    \vee [\exists_{\{x_1,x_2,...,x_p\}}  \ & (\forall_{i\in\{2,...,p\}}     \mathfrak p_\imath(x_{i-1},x_{i}) 
    \leq \alpha) \\ 
   \wedge & (\forall_{i\in \{2,...,p-1\}}   \  M_{\alpha}(x_i)\geq \tau)]
\end{split}
\label{def:connect}
\end{equation}
\end{definition}

The second line denotes direct connectivity between two neighbouring points when $p=2$. The last two lines denote transitive connectivity when $p>2$.

\begin{definition}
A mass-connected cluster $\widetilde C$, which has a mode ${\bf{c}}=\arg\max_{\substack{x\in \widetilde C}}{M}_\alpha(x)$, is a maximal set of points that are mass-connected with its mode, i.e., $\widetilde C=\{x\in D \ | \ MConnect_{\alpha}^{\tau}(x, \bf c)\}$.  
\end{definition}

Note that density-connectivity and density-connected clusters are similarly defined in DBSCAN \cite{ester1996density} when $M_\alpha = \#\{ y \in D\ |\ \mathfrak p_\imath(x,y) \le \alpha\}$ is replaced with $N_\epsilon = \#\{ y \in D\ |\ \ell_p(x,y) \le \epsilon\} $ in the above two definitions. In other words, DBSCAN \cite{ester1996density} which uses $N_\epsilon$ detects density-connected clusters; whereas DBSCAN which uses $M_\alpha$ detects mass-connected clusters.

The only difference between a density-connected cluster and a mass-connected cluster is the dissimilarity measure used in Equation \ref{def:connect}. We called the DBSCAN procedure  which employs $M_\alpha$: MBSCAN, since it detects mass-connected clusters rather than density-connected clusters.

\section{Condition under which MBSCAN detects all mass-connected clusters}
\label{sec_condition}

Let a valley between two cluster modes be the points having the minimum estimated $M_\alpha(\cdot)$, i.e., ${\mathfrak g}_{ij}$, along any path linking cluster modes ${\bf c}_{i}$ and ${\bf c}_{j}$. A path between two points ${\bf x}$ and ${\bf y}$ is non-cyclic linking a sequence of unique points starting with ${\bf x}$ and ending with ${\bf y}$ where adjacent points lie in each other's $\alpha$-neighbourhood: $\mathfrak p_\imath(\cdot,\cdot) \le \alpha$.
 
Because $\mathfrak p_\imath$ (unlike $\ell_2$ used in $N_\epsilon$) is adaptive to the density of local data distribution, it is possible to adjust $\psi$ and $\alpha$ to yield an $M_\alpha$ distribution such that all valley-points have close enough small values, if there exist such $\psi$ and $\alpha$.
 
In other words, for some data distributions $F$,
there exist some $\psi$ and $\alpha$ such that the distribution of $M_\alpha(\cdot)$ satisfies the following condition: 
\begin{equation}
\min_{\substack{k\in \lbrace1,\dots,\aleph \rbrace}} M_\alpha({\bf c}_k) >  \max_{\substack{i\neq j\in \lbrace1,\dots,\aleph \rbrace}} \hat{\mathfrak{g}}_{ij} 
 \label{eqn_condition1}
 \end{equation}
 
\noindent where $\hat{\mathfrak{g}}_{ij}$ is the largest of the minimum estimated $M_\alpha(\cdot)$ along any path linking cluster modes ${\bf c}_{i}$ and ${\bf c}_{j}$.

In data distributions $F$, MBSCAN is able to detect all mass-connected clusters because a threshold $\tau$ can be used to breaks all paths between the modes by assigning regions  with estimated $M_\alpha(\cdot)$ less than ${\tau}$ to noise, i.e., 
\[ \exists_{{\tau}} \forall_{k,i\neq j \in \lbrace 1,...,\aleph \rbrace} M_{{\alpha}}({\bf c}_k) \geqslant {\tau} > \hat{\mathfrak{g}}_{ij} 
\]

\begin{figure}
\centering
\begin{subfigure}{.23\textwidth}
  \centering
  \includegraphics[width=1\linewidth]{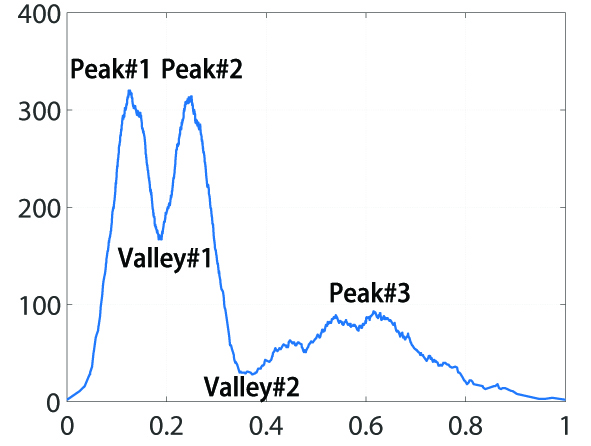}
  \caption{$N_\epsilon$: $\ell_2$}
\end{subfigure}%
\begin{subfigure}{.23\textwidth}
  \centering
  \includegraphics[width=1\linewidth]{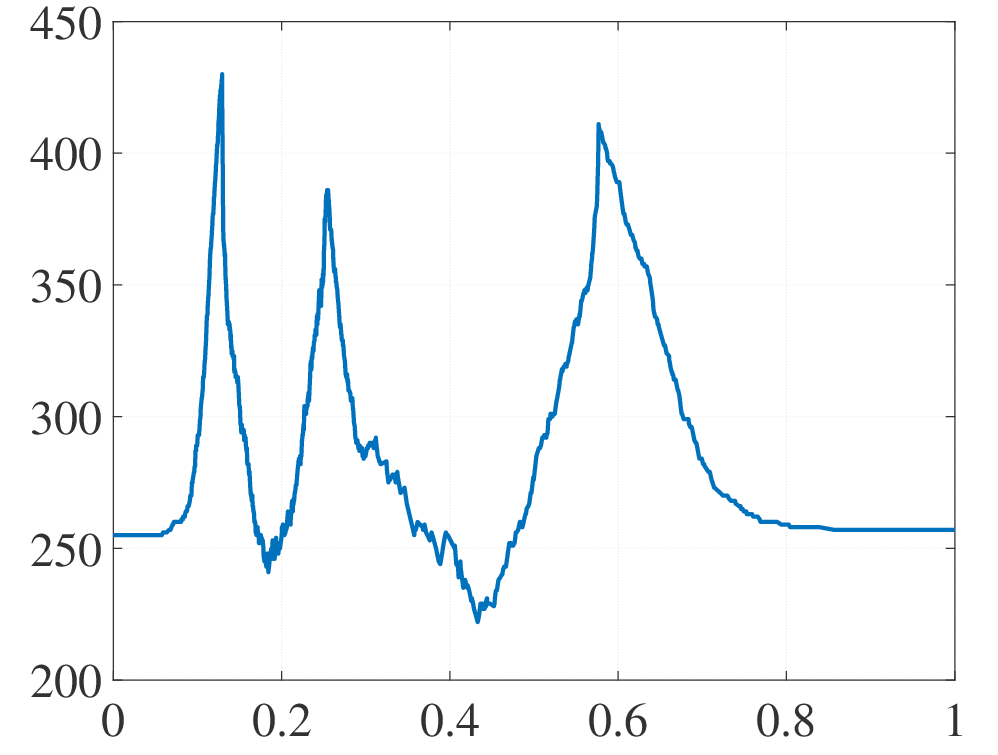}
  \caption{$M_\alpha$: $\mathfrak p_\imath$-aNNE}
\end{subfigure}
\caption{(a) A hard distribution for DBSCAN as estimated by $N_\epsilon$, where DBSCAN (which uses $N_\epsilon$)  fails to detect all clusters using a threshold. (b) The distribution estimated by $M_\alpha$ from the same dataset, where MBSCAN (which uses $M_\alpha$)  succeeds in detecting all clusters using a threshold.}
\label{fig_mass-estimation}
\end{figure}

An example that $F$ subsumes $G$, derived from the same dataset, is shown in Figure \ref{fig_mass-estimation}, where a  hard distribution $G$ in which DBSCAN fails to detect all clusters is shown in Figure \ref{fig_mass-estimation}(a); but MBSCAN succeeds \footnote{Note that the above condition was first described in the context of using mass-based dissimilarity \cite{Ting-MLJ2018}; but not in relation to mass-connected clusters. We have made the relation to mass-connected clusters more explicitly here.}.

In other words, the mass distribution afforded by $M_\alpha$ is more flexible than the density distribution generated by $N_\epsilon$ which leads directly to MBSCAN's enhanced cluster detection capability in comparison with DBSCAN, though both are using exactly the same algorithm, except the dissimilarity.

\begin{figure}
\centering
\begin{subfigure}{0.23\textwidth}
  \centering
  \includegraphics[width=120pt]{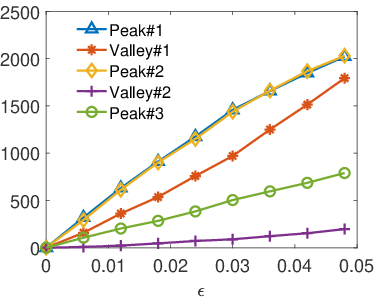}
  \caption{$N_\epsilon$ vs $\epsilon$}
\end{subfigure}
\begin{subfigure}{0.23\textwidth}
  \centering
  \includegraphics[width=120pt]{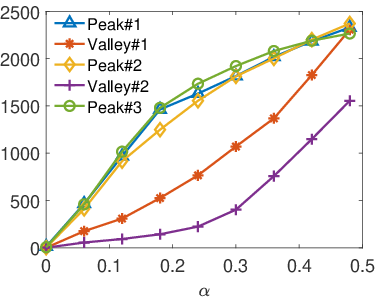}
  \caption{$M_\alpha$ vs $\alpha$}
\end{subfigure}
\caption{Change of neighbourhood density/mass wrt its parameter. $N_\epsilon$ uses $\ell_2$; and $M_\alpha$ uses $\mathfrak p_\imath$-aNNE. The peak numbers and valley numbers refer to those shown in Figure~\ref{fig_mass-estimation}(a). }
\label{fig_change_rate}
\end{figure}

Figure \ref{fig_change_rate} shows the change of neighbourhood function values wrt the change in their parameter for 
$N_\epsilon$ using $\ell_2$ and $M_\alpha$ using $\mathfrak p_\imath$-aNNE. This example shows that no $\epsilon$ exists which enables DBSCAN to detect all three clusters. This is because the line for Peak\#3 (which is the mode of the sparse cluster) has $N_\epsilon$ values in-between those of the two valleys. In contrast, many settings of $\alpha$ of $M_\alpha$ can be used to detect all three clusters because the lines of the two valleys are lower than those of the three peaks. 

\begin{table*}[!htb]
\centering
\caption{Clustering results in $F_1$ scores. The best performer is boldfaced; the second best is underlined. }
\label{tbl_results}
\begin{tabular}{r|rrr||c|ccc|cc}
\hline
\multicolumn{4}{c||}{Datasets}      & DP                & \multicolumn{3}{c|}{DBSCAN}               & \multicolumn{2}{c}{MBSCAN}                    \\ \hline
Name & \#Points & \#Dim. & \#Clusters   & $\ell_2$      & $\ell_2$              & ReScale           & DScale        & iForest                     & aNNE                \\ \hline
\multicolumn{1}{l}{Artificial data} & \multicolumn{3}{r}{(average$\Rightarrow$)} 
                                    & \multicolumn{1}{c}{\textit{0.961}}             & \textit{0.852}                 & \textit{0.941}             & \multicolumn{1}{c}{\textit{0.985}}    & \textit{0.969}                   & \textit{0.981}   \\ \hline
  aggregation & 788 & 2 & 7         & \textbf{1.000}    & 0.997                 & 0.996             & \textbf{1.000}    & 0.996                   & \textbf{1.000}      \\
  compound & 399 & 2 & 6            & 0.867             & 0.791                 & 0.862             & \textbf{0.942}    & 0.875                   & \underline{0.918}   \\
  jain & 373 & 2 & 2                & \textbf{1.000}    & 0.976                 & \textbf{1.000}    & \textbf{1.000}    & \textbf{1.000}          & \textbf{1.000}      \\
  pathbased & 300 & 2 & 3           & 0.943             & 0.828                 & 0.864             & \underline{0.987} & 0.986                   & \textbf{0.995}      \\
  hard distribution & 1500 & 2 & 3  & \underline{0.994} & 0.667                 & 0.985             & \textbf{0.995} & 0.987                   & 0.992               \\ \hline
\multicolumn{1}{l}{High-dimensional data} & \multicolumn{3}{r}{(average$\Rightarrow$)} 
                                    & \multicolumn{1}{c}{\textit{0.627}} & \textit{0.446}                 & \textit{0.521}             & \multicolumn{1}{c}{\textit{0.491}}             & \textit{0.568}                   & \textit{0.727}      \\ \hline
  ALLAML & 72 & 7129 & 2            & 0.706             & 0.484                 & 0.729             & 0.484             & \underline{0.747}       & \textbf{0.820}      \\
  COIL20 & 1440 & 1024 & 20         & 0.724             & 0.842                 & 0.861             & 0.839             & \underline{0.865}       & \textbf{0.952}      \\
  Human Activity & 1492 & 561 & 6   & \textbf{0.595}    & 0.331                 & 0.352             & 0.374             & 0.402                   & \underline{0.502}   \\
  Isolet & 1560 & 617 & 26          & \underline{0.517} & 0.194                 & 0.234             & 0.426             & 0.289                   & \textbf{0.605}      \\
  lung & 203 & 3312 & 5             & \underline{0.703} & 0.489                 & 0.544             & 0.489             & 0.649                   & \textbf{0.921}      \\
  TOX 171 & 171 & 5748 & 4            & \underline{0.519} & 0.336                 & 0.403             & 0.336             & 0.454                   & \textbf{0.563}      \\ \hline
\multicolumn{1}{l}{General data} & \multicolumn{3}{r}{(average$\Rightarrow$)} 
                                    & \multicolumn{1}{c}{\textit{0.876}} & \textit{0.680}                 & \textit{0.820}             & \multicolumn{1}{c}{\textit{0.860}}             & \textit{0.873}                   & \textit{0.896}      \\ \hline
  breast & 699 & 9 & 2              & \textbf{0.970}    & 0.824                 & 0.951             & 0.966             & 0.963                   & \underline{0.964}   \\
  control & 600 & 60 & 6            & 0.736             & 0.531                 & 0.663             & 0.844             & \underline{0.738}       & \textbf{0.854}      \\
  gps & 163 & 6 & 2                 & \underline{0.811} & 0.753                 & \underline{0.811} & \underline{0.811} & \textbf{0.819}          & 0.766               \\
  iris & 150 & 4 & 3                & \underline{0.967} & 0.848                 & 0.905             & 0.926             & 0.966                   & \textbf{0.973}      \\
  seeds & 210 & 7 & 3               & \underline{0.909} & 0.750                 & 0.885             & 0.871             & 0.907                   & \textbf{0.922}      \\
  shape & 160 & 17 & 9              & \underline{0.761} & 0.581                 & 0.680             & 0.722             & 0.725                   & \textbf{0.787}      \\
  thyroid & 215 & 5 & 3             & 0.868             & 0.584                 & 0.850             & 0.828             & \underline{0.915}       & \textbf{0.916}      \\
  WDBC & 569 & 30 & 2               & \textbf{0.933}    & 0.600                 & 0.765             & 0.894             & 0.895                   & \underline{0.927}   \\
  wine & 178 & 13 & 3               & \underline{0.933} & 0.645                 & 0.866             & 0.881             & 0.927                   & \textbf{0.959}      \\ \hline \hline
\multicolumn{4}{r||}{Grand Average}
                                    & 0.823    & 0.653                     & 0.760                 & 0.781                 & 0.805                       & 0.867         \\
\multicolumn{4}{r||}{Number of datasets with the \textbf{Best} $F_1$ score}
                                    & 5    & 0                     & 1                 & 4                 & 2                       & 14         \\
\multicolumn{4}{r||}{\#wins/\#draws/\#loses wrt MBSCAN-$\mathfrak p_\imath$-aNNE}
                                    & 5/2/13            & 0/0/20                & 1/2/18            & 4/2/14            & 1/1/18                  & -                   \\ \hline
\end{tabular}
\end{table*}

\section{Experiments}
\label{sec_experiments}
The aim of the experiments is to compare the clustering performance of DBSCAN using different dissimilarities relative to that of the state-of-the-art density-based clustering algorithm DP \cite{rodriguez2014clustering}. In addition to the three dissimilarity measures, i.e., $\ell_2$, $\mathfrak p_\imath$-iForest and $\mathfrak p_\imath$-aNNE, two recent distance transformation method called ReScale \cite{zhu2016density} and DScale \cite{DSCALE} are also included. Note that DBSCAN using $\mathfrak p_\imath$ are denoted as MBSCAN, as they are mass-based clustering methods.

All algorithms used in our experiments are implemented in Matlab (the source code with demo can be obtained from \url{https://github.com/cswords/anne-dbscan-demo}). 
We produced the GPU accelerated versions of all implementations. The experiments ran on a machine having CPU: i5-8600k 4.30GHz processor, 8GB RAM; and GPU: GTX Titan X with 3072 1075MHz CUDA \cite{4490127} cores \& 12GB graphic memory. 

A total of 20 datasets\footnote{The artificial datasets are from \url{http://cs.uef.fi/sipu/datasets/} \cite{gionis2007clustering,zahn1971graph,chang2008robust,jain2005data} except that the hard distribution dataset is from \url{https://sourceforge.net/p/density-ratio/} \cite{zhu2016density}, 5 high-dimensional data are from \url{http://featureselection.asu.edu/datasets.php} \cite{li2016feature}, and the rest of the datasets are from \url{http://archive.ics.uci.edu/ml} \cite{dua2017uci}.} are used in the experiments. They are from three categories: 5 artificial datasets, 6 high-dimensional datasets, and 9 general datasets. They are selected because they represent diverse datasets in terms of data size, number of dimensions and number of clusters. The data characteristics of these datasets are shown in the first four columns of Table \ref{tbl_results}. 
All datasets are normalised using the $min$-$max$ normalisation so that each attribute is in [0,1] before the experiments begin.

We compared all clustering results in terms of the best $F_{1}$ score \cite{Fmeasure} \footnote{$F_{1}=\frac{1}{k}\sum_{i=1}^{k}\frac{2p_{i}r_{i}}{p_{i}+r_{i}}$, where $p_{i}$ and  $r_{i}$ are the precision and the recall for cluster $i$, respectively. $F_{1}$ is preferred over other evaluation measures such as Purity \cite{Manning:2008} and Normalized Mutual Information (NMI) \cite{strehl2002cluster} because these measures do not take into account noise points which are identified by a clustering algorithm. Based on these measures, algorithms can obtain a high clustering performance by assigning many points to noise, which can be misleading in a comparison.} that is obtained from a search of the algorithm's parameter. 
We search each parameter within a reasonable range. The ranges used for all algorithms/dissimilarities are provided in Table \ref{tbl_param_range}. Because $\mathfrak p_\imath$  used in MBSCAN is based on randomised methods, we report the mean $F_{1}$ score over 10 trials for each dataset. 

\begin{table}[ht]
 
  \renewcommand{\arraystretch}{1.1}
 \setlength{\tabcolsep}{2.8pt}
 	\centering 
 	\caption{Search ranges of parameters used.}
 	\begin{tabular}{@{}c|c|l@{}}
    \hline
 	 & Description & Candidates \\ \hline
 	\multirow{2}{*}{DP} & Target cluster number & $k \in [2...40]$ \\ 
 	 & neighbourhood size in $N_\epsilon$ & $\epsilon \in [0.001...0.999]$  \\ \hline
 	{DBSCAN} & $MinPts$ & $MinPts \in [2...40]$ \\ 
 	 {MBSCAN}& neighbourhood size in $N_\epsilon$ &  $\epsilon \in [0.001...0.999]$ \\ \hline
 	{ReScale} & precision factor & $f=200$ * \\ 
 	{DScale} & neighbourhood size in $N_\eta$ &  $\eta \in [0.05...0.95]$  \\ \hline
 	{aNNE} &  Ensemble size & $t= 200$ \\ 
 	iForest & Subsample size & $\psi \in [2, \lceil n/2 \rceil]$  $\dagger$ \\ \hline
 	\multicolumn{3}{p{8cm}}{* $f$ parameter is required for ReScale only.}\\
 	\multicolumn{3}{p{8cm}}{$\dagger$ A search of 10 values with equal interval in the range.}
 	\end{tabular}
 	\label{tbl_param_range}
\end{table}

\subsection{Clustering results}
Table \ref{tbl_results} shows that MBSCAN using $\mathfrak p_\imath$-aNNE has the best performance overall. Its $F_1$ scores are the best on 14 out of 20 datasets. The closest contender DP has the best $F_1$ scores on 5 datasets only. In two other performance measures, MBSCAN using $\mathfrak p_\imath$-aNNE has 13 wins, 2 draws and 5 losses against DP; and has higher average $F_1$ score too (0.867 versus 0.823). 

One notable standout is on the high-dimensional datasets: MBSCAN using $\mathfrak p_\imath$-aNNE has the largest gap in average $F_1$ score in comparison with other contenders among the three categories of datasets. With reference to DBSCAN, the gap is close to 0.3 $F_1$ score; even compare with the closest contender DP, the gap is 0.1 $F_1$ score. The superiority of  $\mathfrak p_\imath$-aNNE over  $\mathfrak p_\imath$-iForest is also highlighted on these high-dimensional datasets. 

MBSCAN using $\mathfrak p_\imath$-iForest wins over the original DBSCAN on all datasets except one (aggregation). This version of MBSCAN uplifted the clustering performance of DBSCAN significantly to almost the same level of DP. 

\begin{figure}[!htb]
 	\centering
 	\includegraphics[width=\linewidth]{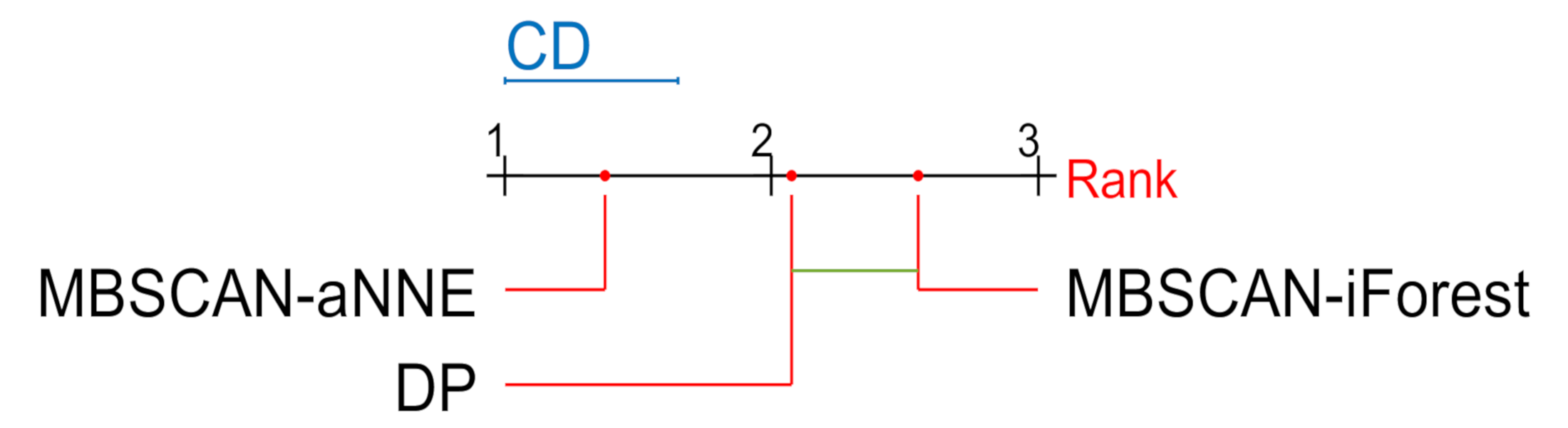}
 	\caption{Critical difference (CD) diagram of the post-hoc Nemenyi test ($\alpha=0.10$). 
 	A line is showed between two algorithms if their rank gap is smaller than CD; otherwise, the difference is significant.
 	}
 	\label{rank}
\end{figure}

A significance test is conducted over MBSCAN with $\mathfrak p_\imath$-aNNE, MBSCAN with $\mathfrak p_\imath$-iForest and DP.
Figure \ref{rank} shows the result of the test---MBSCAN using $\mathfrak p_\imath$-aNNE performs the best and is significantly better than DP and MBSCAN using $\mathfrak p_\imath$-iForest; and there is no significant difference between DP and MBSCAN using $\mathfrak p_\imath$-iForest.

DBSCAN is known to be  sensitive to its parameter settings. As MBSCAN is using exactly the same algorithm, it has same sensitivity. 

A caveat is in order. Although Isolation Similarity consistently outperforms distance measure in DBSCAN, our preliminary experiment using DP shows that the result is mixed. An analysis of DP, similar to that provided in this paper, is required in order to ascertain the condition(s) under which DP performs well and Isolation Similarity can help. 

\subsection{Complexity and runtime}
MBSCAN with $\mathfrak p_\imath$-aNNE is much faster than MBSCAN with $\mathfrak p_\imath$-iForest because the time complexity to build a maximum-size isolation tree and testing one point is $O(\psi^2)$; and aNNE takes $O(\psi)$. 
The space complexity to store trained aNNE model is $O(t \cdot \psi)$; and that of iForest is $O(t \cdot \psi \cdot \log\psi)$.

Table \ref{tbl_runtime} shows the GPU runtime results on the four largest datasets. In contrast to DBSCAN and DP, MBSCAN needs to pre-compute the dissimilarity matrix in the pre-processing step, and this takes $O(n^2)$ time. This pre-processing constitutes the most of the time of MBSCAN reported in Table~\ref{tbl_runtime}. $\mathfrak p_\imath$-aNNE is still faster than ReScale in high-dimensional datasets, though it is one order of magnitude slower than DBSCAN and DP.

\begin{table}[!htb]
\centering
 \renewcommand{\arraystretch}{1.2}
\setlength{\tabcolsep}{5pt}
\caption{Runtime in GPU seconds}
\begin{tabular}{@{}r|c|cc|cc@{}}
\hline
  Datasets & DP & \multicolumn{2}{c|}{DBSCAN} & \multicolumn{2}{c}{MBSCAN}  \\ \cline{3-6}
  & & Original  & ReScale & \multicolumn{1}{c}{iForest} & \multicolumn{1}{c}{aNNE} \\ \hline
 
  Hard dist. & 0.11 & 0.07 & 0.08 & 154 & 0.50 \\
  COIL20 & 0.03 & 0.02 & 3.74 & 762 & 0.45 \\
  Human Act. & 0.10 & 0.03 & 2.12 & 146 &  0.48 \\
  Isolet & 0.08 & 0.03 & 2.64 & 472 & 0.48 \\ \hline
\end{tabular}
\label{tbl_runtime}
\end{table}

\vspace{3mm}
\begin{center}
\textbf{Summary: aNNE versus iForest implementations of Isolation Similarity}
\end{center}

We find that the aNNE implementation of Isolation Similarity is better than the iForest implementation because the aNNE implementation is more:
\begin{itemize}
\item
Effective on datasets with varied densities and high-dimensional datasets.
\item 
Amenable to GPU acceleration because aNNE can be implemented in almost pure matrix manipulations. Thus, aNNE runs many orders of magnitude faster than iForest if a large $\psi$ is required because aNNE in the GPU implementation has almost constant runtime wrt $\psi$.
\item
Stable because aNNE's randomisation is a result of sampling data subsets only.
\end{itemize}

\section{Conclusions}
\label{sec_conclusions}

We make four contributions in this paper:
\renewcommand{\labelenumi}{\arabic{enumi})}
\begin{enumerate}
\item Identifying shortcomings of tree-induced Isolation Similarity; proposing a nearest neighbour-induced Isolation Similarity to overcome these shortcomings; and establishing three advantages of the nearest neighbour-induced Isolation Similarity over the tree-induced one. 

\item Formally proving the characteristic of the  nearest neighbour-induced Isolation Similarity. This is the first proof since the introduction of Isolation Kernel \cite{ting2018IsolationKernel}.

\item Providing a formal definition of mass-connected clusters  and  an  explanation  why  detecting  mass-connected
clusters  is  a  better  approach  in  overcoming  the  shortcoming  of  DBSCAN  (which  detects
density-connected clusters) in datasets with varied densities. This differs fundamentally from the existing density-based
approaches  of  the  original  DBSCAN,  DP  and  ReScale
which all employ a distance measure to compute density.

\item Conducting an empirical evaluation to validate the advantages of (i) nearest-neighbour-induced Isolation Similarity over tree-induced Isolation Similarity; and (ii) mass-based clustering using  Isolation Similarity over four density-based clustering algorithms, i.e., DBSCAN,  DP,  ReScale and DScale.
\end{enumerate}

In addition, we show for the first time that it is possible to uplift the clustering performance of the classic DBSCAN, through the use of nearest-neighbour-induced Isolation Similarity, to surpass that of DP---the state-of-the-art density-based clustering algorithm.

\section*{Acknowledgements}
This material is based upon work supported by eSolutions of Monash University (Xiaoyu Qin);
and partially supported by the Air Force Office of Scientific Research, Asian Office of Aerospace Research and Development (AOARD) under award number: FA2386-17-1-4034 (Kai Ming Ting).

\bibliographystyle{aaai}
\bibliography{AAAI-QinX}

\end{document}